\documentclass[final,twocolumn,3p]{elsarticle}
\usepackage{subcaption}
\usepackage{amsmath,amssymb,amsfonts}
\usepackage{graphicx}
\usepackage{multirow}
\usepackage{booktabs}
\usepackage{xcolor}
\usepackage{colortbl}
\usepackage{arydshln}
\usepackage{hhline}
\usepackage{soul}

\usepackage{algorithmic}
\usepackage{algorithm}
\usepackage{array}
\usepackage{textcomp}
\usepackage{stfloats}
\usepackage{url}
\usepackage{verbatim}
\usepackage{graphicx}
\usepackage{makecell}
\usepackage[colorlinks=True,urlcolor=magenta]{hyperref}

\newcolumntype{s}{>{\columncolor{gray!10}}c}

\journal{Expert Systems with Applications}
\biboptions{authoryear}

\begin{document}

\title{Fully Differentiable Bidirectional Dual-Task Synergistic Learning \\ for Semi-Supervised 3D Medical Image Segmentation}

\author[ets]{Jun Li\corref{cor1}}

\cortext[cor1]{Corresponding author. \\ \textit{Email address:} dirk.li@outlook.com.}
\address[ets]{School of Electrical Engineering, Southwest Jiaotong University, Chengdu, 611756, China.}

%%%%%%%%%%%%%%%%%%%%%%%%%%%%%%%%%%%%%%%%%%%%%%%%%%%%%%%%%%%%%%%%%%%%%%%%
\begin{frontmatter}
\begin{abstract}

Semi-supervised learning relaxes the need of large pixel-wise labeled datasets for image segmentation by leveraging unlabeled data. The scarcity of high-quality labeled data remains a major challenge in medical image analysis due to the high annotation costs and the need for specialized clinical expertise. Semi-supervised learning has demonstrated significant potential in addressing this bottleneck, with pseudo-labeling and consistency regularization emerging as two predominant paradigms. Dual-task collaborative learning, an emerging consistency-aware paradigm, seeks to derive supplementary supervision by establishing prediction consistency between related tasks. However, current methodologies are limited to unidirectional interaction mechanisms (typically regression-to-segmentation), as segmentation results can only be transformed into regression outputs in an offline manner, thereby failing to fully exploit the potential benefits of online bidirectional cross-task collaboration. Thus, we propose a fully \textbf{D}ifferentiable \textbf{Bi}directional \textbf{S}ynergistic \textbf{L}earning (DBiSL) framework, which seamlessly integrates and enhances four critical SSL components: supervised learning, consistency regularization, pseudo-supervised learning, and uncertainty estimation. Experiments on two benchmark datasets demonstrate our method's state-of-the-art performance. Beyond technical contributions, this work provides new insights into unified SSL framework design and establishes a new architectural foundation for dual-task-driven SSL, while offering a generic multitask learning framework applicable to broader computer vision applications. The code will be released at \url{https://github.com/DirkLiii/DBiSL}.

\end{abstract}


\end{frontmatter}
% \setlength{\parskip}{3pt}

%%%%%%%%%%%%%%%%%%%%%%%%%%%%%%%%%%%%%%%%%%%%%%%%%%%%%%%%%%%%%%%%%%%%%%%%
\section{Introduction}

Deep learning, predicated on data-driven paradigms, has achieved unprecedented performance across a wide spectrum of computer vision tasks. However, the empirical success of these models is critically contingent upon the scale, diversity, and annotation quality of the training data. This fundamental data dependency presents a significant bottleneck, posing substantial challenges for numerous tasks, particularly in scenarios characterized by a paucity of readily available labeled data \cite{li2022dual}. This issue is especially pronounced in the medical image community, where annotation demands specialized clinical expertise and physician involvement \cite{WU2025103547,li2023generalizable}. Furthermore, the typical annotation granularity required, such as pixel-wise segmentation or slice-by-slice delineation across volumetric data, is exceedingly labor-intensive and time-consuming, frequently demanding several hours per individual sample \cite{isensee2021nnu}. This not only imposes considerable strain on already limited healthcare resources but also significantly impedes the widespread clinical deployment and scalability of deep learning-based medical image analysis tools.

%%%%%%%%%%%%%%%%%%%%%%%%%%%%%%%%%%%%%%%%%%%%%%%%%%%%%%%%%%%%%%%%%%%%%%%%
\begin{figure}[t]
\centering
\includegraphics[width=0.98\columnwidth]{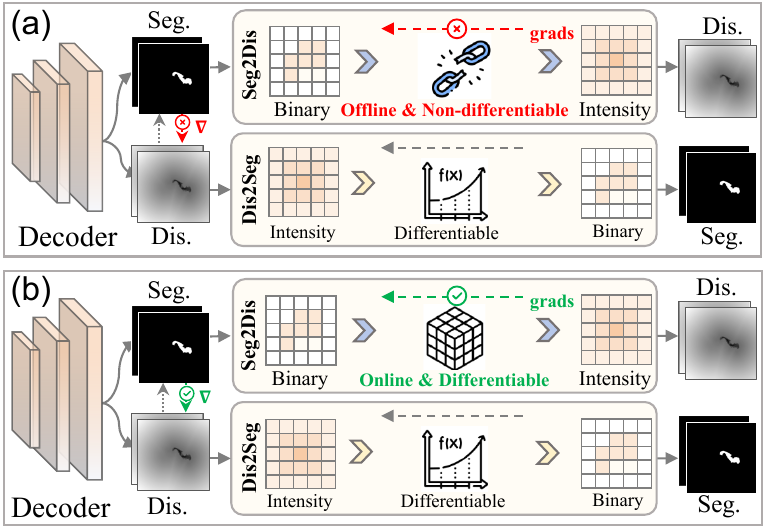}
\caption{Different dual-task structures. (a) Unidirectional: Information flows one-way, from one task to another only. (b) Bidirectional (Proposed): Bidirectional task interaction, information can flow seamlessly between two tasks while preserving gradient continuity.}
\label{fig:fig1}
\end{figure}
%%%%%%%%%%%%%%%%%%%%%%%%%%%%%%%%%%%%%%%%%%%%%%%%%%%%%%%%%%%%%%%%%%%%%%%%

To address the challenges of data scarcity and high annotation costs, semi-supervised learning (SSL) has emerged as a compelling paradigm. It can effectively integrate both limited labeled data and readily available abundant unlabeled data into the training process, and has achieved notable success in medical image segmentation community \cite{yang2025unimatch}. Generally, most existing SSL methods are structured around two principal components: a supervised learning branch that utilizes the available labeled data, and an unsupervised learning branch that exploits the potential within the unlabeled data. The supervised component serves to establish a reliable foundation and enhance model robustness by training on ground truth labels, while the unsupervised component is designed to leverage the inherent structure, consistency, or other valuable information present in the vast pool of unlabeled data to significantly improve overall performance \cite{jiao2024learning}. 

Reflecting the complexity and potential of harnessing unlabeled information, current SSL research has predominantly emphasized the refinement of the unsupervised learning component. This involves developing complex techniques to extract and discern beneficial supervisory signals from unlabeled data. Prevailing strategies include generating pseudo-labels for unlabeled samples, enforcing prediction consistency under various data augmentations or model perturbations, utilizing model uncertainty to identify reliable learning targets, or employing synergistic combinations of these techniques \cite{10771804}.

Nevertheless, existing methods \cite{luo2022semi,wu2023exploring} often underemphasize the role of labeled data within the SSL context. The learning efficacy from labeled data, however, significantly shapes model optimization trajectories. Consequently, standard supervised training alone may not fully exploit the potential of labeled data. Moreover, while there is a trend towards integrating diverse SSL techniques for richer supervision, some approaches simply aggregate methods via parallel architectures. This may result in suboptimal information exchange, or even functional independence, between technical branches, thereby diminishing the effectiveness of the integrated strategy.

To mitigate these limitations, recent SSL studies have investigated distance regression—a task intrinsically related to segmentation, predicting the minimal distance from each pixel to the object boundary—as an auxiliary task \cite{luo2021semi,li2020shape}. The objective is to enhance the integration of supervised and unsupervised signals through the synergistic interaction of these tasks. However, these methods often resort to offline methods to transform segmentation probability maps into distance prediction maps (Figure \ref{fig:fig1}) \cite{ma2020distance}. This results in a unidirectional information flow—from regression to segmentation—hindering the establishment of a bidirectional task association. Therefore, this work aims to introduce a fully differentiable bidirectional task transformer, directly establishing gradient associations between pixel segmentation and distance regression tasks during training, thereby achieving their collaborative optimization.

To this end, we propose a \textbf{D}ifferentiable \textbf{Bi}directional \textbf{S}ynergistic \textbf{L}earning (DBiSL) framework. DBiSL employs a differentiable 3D convolution operator to emulate offline distance map generation, maintaining complete gradient flow while enabling task inter-conversion. Our core idea lies in this differentiable bidirectional transformer, designed to construct a concise yet efficient SSL system. Ultimately, DBiSL aims to synergistically enhance various SSL components, including supervised learning, consistency regularization, pseudo-labeling, and uncertainty estimation, within a global and consistent objective. Specifically, supervised learning incorporates standard segmentation supervision on labeled data, as well as cross-task supervision. The latter, facilitated by the differentiable task transformer, realizes the supervision of pixel segmentation tasks using distance regression annotations, and vice versa. Consistency regularization is predicated on the assumption that task outputs should be consistent, enforced in both pixel and distance spaces via the bidirectional transformer. Pseudo-labeling and the corresponding uncertainty estimation modules are implemented by integrating task outputs from different branches. Their primary objective is to filter high-confidence regions for pseudo-label supervision and minimize noise interference. 

We summarize our contributions: \textbf{(i)} a fully differentiable bidirectional task transformer, enabling gradient-propagatable inter-task interaction; \textbf{(ii)} a comprehensive SSL framework, which effectively strengthens all constraint losses through a streamlined and efficient design; \textbf{(iii)} a flexible technical architecture, offering strong compatibility with mainstream SSL techniques and notable adaptability to other vision tasks.

%%%%%%%%%%%%%%%%%%%%%%%%%%%%%%%%%%%%%%%%%%%%%%%%%%%%%%%%%%%%%%%%%%%%%%%%
\begin{figure*}[t]
\includegraphics[width=0.99\linewidth]{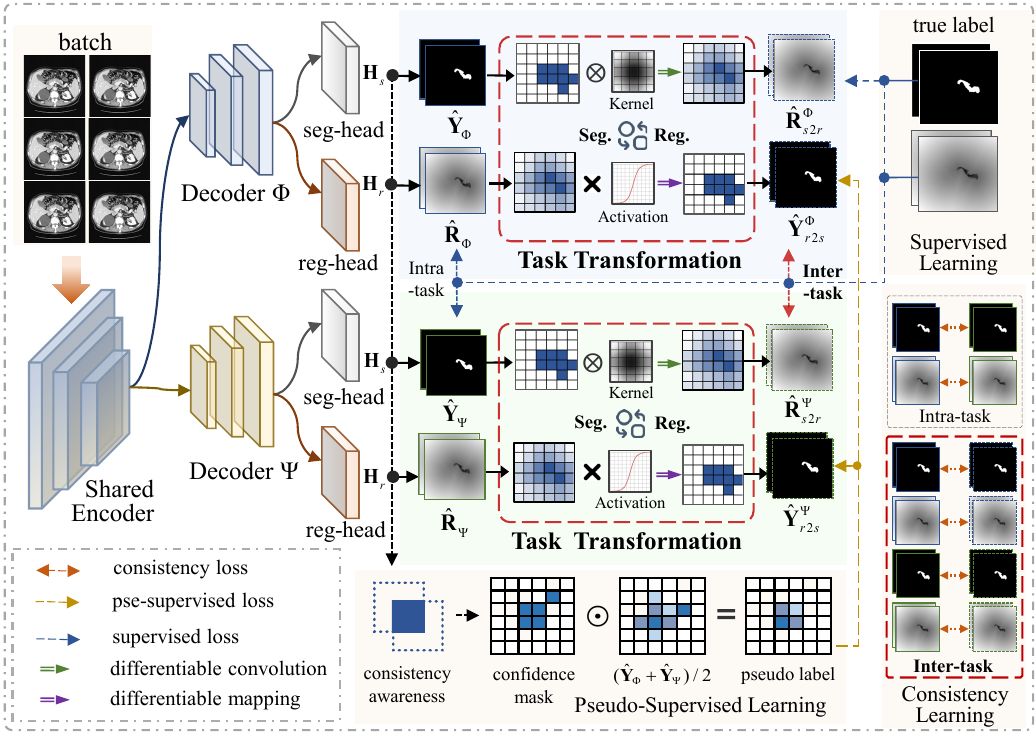}
\centering
\caption{Overview of the DBiSL framework. A shared encoder with segmentation and distance-regression heads is coupled by our fully differentiable bidirectional task transformer, enabling online dual-task interaction and supporting cross-task supervision and cross-task consistency within a unified SSL pipeline.}
\label{fig:fig2}
\end{figure*}
%%%%%%%%%%%%%%%%%%%%%%%%%%%%%%%%%%%%%%%%%%%%%%%%%%%%%%%%%%%%%%%%%%%%%%%%

%%%%%%%%%%%%%%%%%%%%%%%%%%%%%%%%%%%%%%%%%%%%%%%%%%%%%%%%%%%%%%%%%%%%%%%%
\section{Related Work}
%%%%%%%%%%%%%%%%%%%%%%%%%%%%%%%%%%%%%%%%%%%%%%%%%%%%%%%%%%%%%%%%%%%%%%%%
\label{sec:related}

\subsection{Semi-Supervised Medical Image Segmentation}

Semi-supervised learning has emerged as a promising approach to alleviate the substantial dependence of deep learning models on extensive labeled datasets, offering a promising avenue for scenarios where data annotation is costly or scarce \cite{HAN2024123052}. Existing methods can be broadly categorized into three categories: pseudo-labeling \cite{shi2021inconsistency,han2022effective}, consistency regularization \cite{ZHU2024102383,luo2021semi,bai2023bidirectional,wang2021self}, and hybrid approaches that combine elements of both \cite{wang2022semi,wang2023mcf}. Pseudo-labeling methods \cite{shi2021inconsistency,han2022effective} utilize high-confidence predictions on unlabeled data as pseudo-labels, incorporating them into the training pipeline. Related researches primarily focuses on developing strategies to improve the reliability and quality of the generated pseudo-labels, often employing techniques such as ensemble methods, iterative self-training pipelines, or filtering based on uncertainty. Consistency regularization approaches \cite{ZHU2024102383,luo2021semi,bai2023bidirectional,wang2021self,wang2024semi}, on the other hand, enhance model robustness by enforcing consistent predictions under perturbations at data, feature, network, and task levels, and may incorporate GANs \cite{10486949,hou2022semi} or contrastive learning \cite{zhao2022cross} for better representations. Recognizing the distinct yet complementary strengths of pseudo-labeling and consistency regularization, hybrid methods \cite{wang2022semi,wang2023mcf} aim to synergistically integrate elements from both categories. For instance, uncertainty-aware frameworks can fuse consistency learning with uncertainty-guided pseudo-label refinement strategies \cite{yu2019uncertainty}. While hybrid methods often yield significant performance gains, the forced integration of multiple technical pathways may introduce additional complexity to model design and the training process. Thus, this work aims to construct a concise yet efficient framework, based on our differentiable bidirectional task transformer, capable of naturally integrating the aforementioned SSL techniques without compromising training stability. 

\subsection{Multi-task-based Medical Image Analysis}

Given that medical image analysis often necessitates performing multiple interrelated tasks on the same input data, multi-task learning has emerged as a compelling and efficient paradigm. Existing multi-task architectures developed for medical image analysis can typically be categorized into four main types based on their inter-task operational modes: cascaded \cite{hamghalam2020high,wang2020automatic}, parallel \cite{li2023semi,yang2021hybrid}, interactive \cite{zhou2020benchmark,zhou2020hi}, and hybrid \cite{tomar2021self,tang2019nodulenet} architectures. Cascaded architectures process tasks in sequence, using the output of one task as input for the next, common in applications like tumor imaging with pre-training image synthesis \cite{hamghalam2020high}. Parallel architectures, conversely, employ independent pathways for each task, typically sharing a feature encoder to enhance feature representation and segmentation accuracy, as exemplified by Aakarsh et al.'s four parallel decoders for segmentation and classification \cite{malhotra2022multi}. Interactive architectures facilitate information exchange between tasks to leverage shared knowledge; for instance, \cite{xu2021asymmetric} linked bladder and rectum segmentation to improve performance in complex scenarios. Hybrid architectures aim to integrate elements of the above approaches to achieve task-specific information fusion. While effective, they often rely on large-scale annotated datasets. Besides, the potential competition between tasks may misalign optimization, prioritizing auxiliary over primary tasks. Thus, we aim to realize bidirectional task transformation and deep interaction by fostering alignment among optimization criteria, while preserving the integrity of gradient information.

%%%%%%%%%%%%%%%%%%%%%%%%%%%%%%%%%%%%%%%%%%%%%%%%%%%%%%%%%%%%%%%%%%%%%%%%
\section{Method}
\label{sec:methods}

We propose a fully differentiable bidirectional task transformer and a dual-task synergistic learning framework, DBiSL, as shown in Figure \ref{fig:fig2}. The framework comprises a shared encoder $\mathbf{E}$ and two parallel decoders, $\mathbf{D}\Phi$ and $\mathbf{D}\Psi$, each equipped with segmentation and regression heads, $\mathbf{H}_s$ and $\mathbf{H}_r$, to produce pixel-wise segmentation $\hat{\mathbf{Y}}$ and distance regression outputs $\hat{\mathbf{R}}$. Supervised regression labels are generated online and in a patch-based manner from segmentation labels using our task transformer, distinguishing our approach from prior work \cite{luo2021semi,ma2020distance,hu2020automatic,wang2020deep}; further details are provided in Section \ref{sec:results}. Through bidirectional task transformation and interaction, DBiSL flexibly incorporates a range of mainstream SSL techniques, yielding a hybrid model with aligned objectives. Algorithm \ref{alg1} outlines the workflow. The framework leverages our proposed differentiable task transformer and incorporates popular paradigms in semi-supervised learning. It is structured around three core modules: supervised learning, consistency regularization \cite{luo2021semi,bai2023bidirectional,wang2021self}, and pseudo-supervised learning \cite{shi2021inconsistency,han2022effective}. We subsequently elaborate on how each component is effectively aligned with the proposed transformer, demonstrating their compatibility and synergistic integration.

\subsection{Backbone Network}
Although our method is model-agnostic and can be seamlessly incorporated into a broad range of segmentation backbones, we adopt the standard V-Net \cite{milletari2016v} architecture for fair comparisons with prior semi-supervised methods. The network comprises a single encoder and two decoders, each equipped with dropout and batch normalization. The two decoders employ distinct upsampling strategies and are attached to two task-specific prediction heads, allowing them to process the shared encoded representation independently while jointly producing segmentation and regression outputs. 

\subsection{Differentiable Bidirectional Transformation}
\subsubsection{Regression to segmentation}

There exists a strict spatial correspondence between distance maps and segmentation probability maps. The distance value for each pixel in the distance regression output can be interpreted as a measure of segmentation confidence for that pixel, specifically reflecting the degree to which the pixel's segmentation probability deviates from 0.5 \cite{luo2021semi}. When the distance regression task determines that a pixel is closer to the object boundary, the model exhibits relatively lower segmentation confidence for that pixel in the corresponding segmentation task. This is manifested as a segmentation probability that is closer to 0.5. Based on this observation, a direct conversion from regression results can be formulated as follows:

\begin{equation}
\label{eq1}
    \hat{\textbf{Y}}_{r2s}=\textbf{T}_{r2s}(\hat{\textbf{R}})=\frac{1}{1+e^{-k \cdot \hat{\textbf{R}}}}.
\end{equation}

\subsubsection{Segmentation to regression}

A key bottleneck in converting segmentation probability maps to distance maps lies in implementing two sequential transformations: probability maps to binary masks, and then binary masks to distance maps \cite{ma2020distance,hu2020automatic,wang2020deep}. While these transformations are readily performed in offline settings, the requirement for continuous gradient backpropagation within model training paradigms introduces substantial complexity. For the conversion of probability maps to binary masks, readily available built-in functions within modern machine learning frameworks offer convenient solutions. However, a definitive and differentiable solution for transforming binary masks to distance maps, especially for volumetric medical images, remains an open challenge. Inspired by \cite{riba2020kornia}, we propose a fully differentiable 3D transformation method that approximates the distance transform via convolution operations, employing a custom-designed kernel $\textbf{K}$ to iteratively refine the precision of distance approximation: 

\begin{equation}
    \textbf{K}(x,y,z)=\exp(-\frac{\sqrt{x^2+y^2+z^2}}{h}).
\end{equation}
where $(x, y, z)$ represents the relative spatial coordinates within the kernel, and $h$ is a hyperparameter controlling approximation quality. To generate the signed distance map, we perform two distinct distance transform computations: one on the mask and another on its inverse (1 - mask), quantifying interior/exterior distances to the boundary:

\begin{equation}
    \textbf{S}_{in}=\textbf{DT}(\hat{\textbf{Y}}, \textbf{K}), \quad
    \textbf{S}_{out}=\textbf{DT}(1-\hat{\textbf{Y}}, \textbf{K}).
\end{equation}
where $\textbf{DT}$ is our proposed transformer. Subsequently, these two distance maps are normalized and subtracted to yield the signed distance map \cite{ma2020distance}, which encodes both the distance to the nearest boundary and directional information:

\begin{equation}
    \hat{\textbf{R}}_{s2r}=\frac{\textbf{S}_{in} - \min(\textbf{S}_{in})}{\max(\textbf{S}_{in}) - \min(\textbf{S}_{in})}-\frac{\textbf{S}_{out} - \min(\textbf{S}_{out})}{\max(\textbf{S}_{out}) - \min(\textbf{S}_{out})}.
\end{equation}
where $\min$ and $\max$ are the maximum and minimum functions, respectively. Thus, the transformer from the probability map to the distance map can be simply expressed as:

\begin{equation}
\label{eq5}
    \hat{\textbf{R}}_{s2r}=\textbf{T}_{s2r}(\hat{\textbf{Y}}, \textbf{K}).
\end{equation}

\begin{algorithm}[!t]
    \caption{The Differentiable Bidirectional Synergistic Learning framework}
    \label{alg1}
    \begin{algorithmic}
    \STATE {\textbf{Input}} 
    \STATE our model $\textbf{F}_{DBiSL}$, labeled data $\textbf{X}^{l}$, unlabeled data $\textbf{X}^{u}$
    \STATE {\textbf{Output}} 
    \STATE trained model $\textbf{F}_{DBiSL}$
    \STATE {\textbf{Training}} 
    \STATE \textbf{for} iteration $\leftarrow 0 $ to max\_iteration \textbf{do}
    \STATE \hspace{0.5cm} select $\textbf{X}^{l}$ and $\textbf{X}^{u}$ from random mini-batches
    \STATE \hspace{0.5cm} get outputs $\hat{\textbf{Y}}_{\{\Phi, \Psi\}}$, $\hat{\textbf{R}}_{\{\Phi, \Psi\}}$ from all heads
    \STATE \hspace{0.5cm} get transformed outputs $\hat{\textbf{Y}}_{r2s}$, $\hat{\textbf{R}}_{s2r}$  
    \STATE \hspace{0.5cm} \textbf{Supervised learning}
    \STATE \hspace{0.5cm} build intra-task supervision $\textbf{L}_{sup}^{it}$
    \STATE \hspace{0.5cm} build inter-task supervision $\textbf{L}_{sup}^{ct}$
    \STATE \hspace{0.5cm} \textbf{Consistency learning}
    \STATE \hspace{0.5cm} Construct intra-task consistency $\textbf{L}_{con}^{it}$
    \STATE \hspace{0.5cm} Construct inter-task consistency $\textbf{L}_{con}^{ct}$
    \STATE \hspace{0.5cm} \textbf{Pseudo-supervised learning}
    \STATE \hspace{0.5cm} generate pseudo label $\overline{\textbf{Y}}$
    \STATE \hspace{0.5cm} generate confidence mask $\textbf{M}$
    \STATE \hspace{0.5cm} build pseudo-supervision $\textbf{L}_{pse}$ 
    \STATE \hspace{0.5cm} optimize model with weighted loss $\textbf{L}_{all}$
    \STATE \textbf{return} $\textbf{F}_{DBiSL}$
\end{algorithmic}
\end{algorithm}
\setlength{\textfloatsep}{10pt}

\subsection{Supervised Learning}

\subsubsection{Intra-task supervised learning}

In SSL scenarios, supervised learning is predominantly driven by limited labeled data and often relies on standard supervision strategies, potentially underutilizing rich geometric information within annotations \cite{jiao2024learning,HAN2024123052}. To this end, we introduce a bidirectional cross-task supervision mechanism that leverages dense information from distance regression (pixel segmentation) labels to supervise the complementary pixel segmentation (distance regression) task. Concretely, let $\textbf{X}^l$ and $\textbf{Y}$ denote 3D medical images and segmentation annotations, the segmentation task is supervised by a weighted sum of Dice loss and cross-entropy loss, denoted as $\mathcal{L}_{seg}$. Thus, the normal intra-task supervision loss is:

\begin{equation}
\label{eq6}
\textbf{L}_{sup}^{it}= \sum_{\omega \in \{\Phi, \Psi\}} (\mathcal{L}_{seg}(\hat{\textbf{Y}}_\omega,\textbf{Y})+\mathcal{L}_{mse}(\hat{\textbf{R}}_\omega,\textbf{R})).
\end{equation}

\subsubsection{Inter-task supervised learning}
The distance regression labels $\textbf{R}$ are generated in real-time from $\textbf{Y}$ via our transformer $\textbf{T}$. After normalization to [-1, 1], negative values in $\textbf{R}$ pixels inside the object contour, and positive values outside. The distance regression task is supervised using the mean squared error loss $\mathcal{L}_{mse}$:  

\begin{equation}
\label{eq7}
\begin{split}
\textbf{L}_{sup}^{ct}= & \sum_{\omega \in \{\Phi, \Psi\}} (\mathcal{L}_{mse}(\textbf{T}_{s2r}(\hat{\textbf{Y}}_\omega),\textbf{R}) \\ & +\mathcal{L}_{seg}(\textbf{T}_{r2s}(\hat{\textbf{R}}_\omega),\textbf{Y})).
\end{split}
\end{equation}
where $s2r$ ($r2s$) means the transformation from segmentation (regression) to regression (segmentation). Thus, the supervision loss can be calculated with:

\begin{equation}
\textbf{L}_{sup}= \textbf{L}_{sup}^{it} + \textbf{L}_{sup}^{ct}.
\end{equation}

\subsection{Consistency Learning}

Consistency learning has emerged as a prevalent technique within SSL paradigms \cite{HAN2024123052,jiao2024learning}. The fundamental requirement for continuous gradient propagation during training often restricts current related methods to implementing unidirectional task conversion, thereby impeding the realization of fully bidirectional inter-task interactions. We address this by leveraging our differentiable bidirectional transformer to formulate dual consistency regularization, covering both intra-task and inter-task scenarios.

\subsubsection{Intra-task consistency learning}
Intra-task consistency learning is based on the principle that a robust model, trained under homogeneous paradigms, should demonstrate strong agreement across its task-specific branches. In line with this, we develop an intra-task consistency mechanism to encourage greater consistency between segmentation (regression) outputs from different branches within the same task:

\begin{equation}
\label{eq9}
\textbf{L}_{con}^{it}= \mathcal{L}_{mse}(\hat{\textbf{Y}}_\Phi,\hat{\textbf{Y}}_\Psi)+\mathcal{L}_{mse}(\hat{\textbf{R}}_\Phi,\hat{\textbf{R}}_\Psi).
\end{equation}

\subsubsection{Inter-task consistency learning}
Cross-task consistency learning is grounded in a distinct hypothesis: an intrinsic correlation exists between pixel segmentation and distance regression tasks.  Therefore, the outputs of the model for these two tasks are expected to demonstrate inherent self-consistency, rather than exhibiting mutual discrepancies. To enforce this principle, we construct an inter-task consistency loss function with transformer $\textbf{T}$:

\begin{equation}
\label{eq10}
\begin{split}
\textbf{L}_{con}^{ct} = & \sum_{\omega \in \{\Phi, \Psi\}} \big( \mathcal{L}_{mse}(\textbf{T}_{s2r}(\hat{\textbf{Y}}_\omega),\hat{\textbf{R}}_\omega) \\
& + \mathcal{L}_{mse}(\textbf{T}_{r2s}(\hat{\textbf{R}}_\omega),\hat{\textbf{Y}}_\omega) \big).
\end{split}
\end{equation}
Thus, the consistency learning loss is denoted as:

\begin{equation}
\textbf{L}_{con}= \textbf{L}_{con}^{it} + \textbf{L}_{con}^{ct}.
\end{equation}

\subsection{Pseudo-supervised Learning}

\subsubsection{Pseudo-label generation}

Effectively, the aggregation of outputs from dual-branch, dual-task architecture constitutes a simplified form of ensemble learning. This approximation effectively mitigates instabilities stemming from model initialization, distributional data biases, or intrinsic architectural constraints, thus synthesizing more robust and dependable pseudo-supervisory signals \cite{10771804}. Therefore, pseudo-labels $\overline{\textbf{Y}}$ are derived by averaging the segmentation predictions:

\begin{equation}
\label{eq12}
    \overline{\textbf{Y}} = \frac{1}{2}(\hat{\textbf{Y}}_\Phi + \hat{\textbf{Y}}_\Psi).
\end{equation}

\subsubsection{Uncertainty estimation}
The uncertainty estimation is performed jointly with the four results from the predictive outputs of both tasks across the dual branches. Specifically, this study employs a conservative voting strategy predicated on a "unanimous consent" principle.  Under this principle, a pixel prediction is deemed reliable exclusively when the predictive outcomes from all tasks exhibit complete agreement. The construction of the confidence mask is formally represented as: 

\begin{equation}
\label{eq13}
    \textbf{M} = \mathbb{I}(\Xi(\hat{\textbf{Y}}_\Phi) == \Xi(\hat{\textbf{Y}}_\Psi) == \Xi(\hat{\textbf{R}}_\Phi) == \Xi(\hat{\textbf{R}}_\Psi)).
\end{equation}
where $\textbf{M}$ denotes the confidence mask, $==$ denotes the equality comparison, $\mathbb{I}(\cdot)$ and $\Xi$ are the indicator function and threshold function, respectively. In summary, the pseudo-supervised learning loss can be summarized as:
\begin{equation}
\label{eq14}
    \textbf{L}_{pse} = \mathcal{L}_{ce}(\hat{\textbf{Y}}_\Phi, \overline{\textbf{Y}}) \odot \textbf{M} + \mathcal{L}_{ce}(\hat{\textbf{Y}}_\Psi, \overline{\textbf{Y}}) \odot \textbf{M}.
\end{equation}
where $\odot$ indicates the element-wise product. The overall loss can be summarized as:
\begin{equation}
\label{eq15}
    \textbf{L}_{all} = \lambda(\textbf{L}_{sup} + \textbf{L}_{con})+\beta\textbf{L}_{pse}.
\end{equation}
where $\lambda$ and $\beta$ are weights that balance different losses. $\lambda$ is assigned an empirical value of 0.5 and $\beta$ is scheduled to increase.

\begin{table*}[ht]
\centering
\caption{Comparisons with semi-supervised segmentation methods on LA dataset. Note: We only compare methods reporting results from the \textbf{final training checkpoint} (See Sec.\ref{sec:csd}).}
\begin{tabular}{lccccc}
\toprule
\multirow{2}{*}{Method} & \multicolumn{2}{c}{Scans used} & \multicolumn{3}{c}{Metrics} \\
\cmidrule(lr){2-3} \cmidrule(lr){4-6} 
& Labeled & Unlabeled & Dice (\%) $\uparrow$ & ASD (mm) $\downarrow$ & 95HD (mm) $\downarrow$ \\
\midrule
Supervised baseline & 16 & 0 & 82.36 & 3.08 & 11.87 \\
Supervised baseline & 80 & 0 & 91.88 & 1.37 & 4.60 \\
\midrule
TAC \cite{chen2022semi} & 16 & 64 & 87.75 & 2.04 & 9.45 \\
AUSS \cite{adiga2024anatomically} & 16 & 64 & 88.60 & / & 7.61  \\
MCF \cite{wang2023mcf} & 16 & 64 & 88.71 & 1.90 & 6.32  \\
UA-MT \cite{yu2019uncertainty} & 16 & 64 & 88.88 & 2.26 & 7.32 \\
DTC \cite{luo2021semi} & 16 & 64 & 89.42 & 2.10 & 7.32  \\
SASSNet \cite{li2020shape} & 16 & 64 & 89.54 & 2.20 & 8.24  \\
DUWM \cite{wang2020double} & 16 & 64 & 89.65 & 2.03 & 7.04 \\
BaPC \cite{wang2024boundary} & 16 & 64 & 89.71 & 1.85 & 6.08  \\
SCC \cite{liu2022contrastive} & 16 & 64 & 89.81 & 1.82 & 7.15 \\
\midrule
DBiSL & 16 & 64 & \textbf{90.54} & \textbf{1.80} & \textbf{6.05} \\
\bottomrule
\end{tabular}
\label{tab:tab1}
\end{table*}

\section{Results}
\label{sec:results}
\subsection{Datasets}

We performed comprehensive evaluations on three public benchmark datasets: the \textbf{Left Atrium (LA)} dataset \cite{XIONG2021101832}, the \textbf{Pancreas-CT} dataset \cite{roth2015deeporgan} and \textbf{Brain Tumor Segmentation Challenge (BraTS2019)} dataset \cite{menze2014multimodal}. The LA dataset consists of 100 gadolinium-enhanced 3D cardiac MRI scans, originally acquired at an isotropic resolution of 0.625 × 0.625 × 0.625 mm. The Pancreas-CT dataset comprises 82 contrast-enhanced abdominal CT scans, with an in-plane resolution of 512×512 mm and slice thicknesses ranging from 1.5 to 2.5 mm. The BraTS2019 dataset comprises multi-parametric brain MRI volumes from 335 glioma subjects, where each case includes four sequences (T1, T1ce, T2, and FLAIR) and corresponding annotations for three tumor-related regions (enhancing tumor, tumor core, and whole tumor). To ensure a consistent and fair protocol, we follow the evaluation setting commonly used in prior studies \cite{luo2022semi,xu2023ambiguity,miao2023caussl,su2024mutual,9741294,zhang2023uncertainty} and focus on WT segmentation from FLAIR, as WT delineation captures the overall disease extent and is clinically relevant for surgical planning, particularly in low-grade gliomas. We adopt the same preprocessing and split strategy as previous works: 250 volumes for training, 25 for validation, and 60 for testing. All images were uniformly resampled to an isotropic resolution of 1.0 × 1.0 × 1.0 mm. Data preprocessing protocols and the stratification of labeled and unlabeled data were consistently maintained with prior studies, which is recognized as the canonical experimental setup.

\begin{table*}[t]
\centering
\caption{Comparisons with semi-supervised segmentation methods on Pancreas-CT dataset. Note: We only compare methods reporting results from the \textbf{final training checkpoint} (See Sec.\ref{sec:csd}).}
\begin{tabular}{lccccc}
\toprule
\multirow{2}{*}{Method} & \multicolumn{2}{c}{Scans used} & \multicolumn{3}{c}{Metrics} \\
\cmidrule(lr){2-3} \cmidrule(lr){4-6} 
& Labeled & Unlabeled & Dice (\%) $\uparrow$ & ASD (mm) $\downarrow$ & 95HD (mm) $\downarrow$ \\
\midrule
Supervised baseline & 12 & 0 & 65.60 & 1.87 & 12.83 \\
Supervised baseline & 62 & 0 & 81.69 & 1.18 & 6.00 \\
\midrule
MCF \cite{wang2023mcf} & 12 & 50 & 75.00 & 3.27 & 11.59 \\
BaPC \cite{wang2024boundary} & 12 & 50 & 76.36 & 2.83 & 10.59 \\
DTC \cite{luo2021semi} & 12 & 50 & 78.27 & 2.25 & 8.36 \\
CoraNet \cite{shi2021inconsistency} & 12 & 50 & 79.67 & 1.89 & 7.59 \\
AUA \cite{wu2023exploring} & 12 & 48 & 79.81 & 1.64 & 5.90 \\
URPC \cite{luo2022semi} & 12 & 50 & 80.31 & \textbf{1.43} & \textbf{4.39} \\
CauSSL \cite{miao2023caussl} & 12 & 50 & 80.92 & 1.53 & 8.11 \\
\midrule
DBiSL & 12 & 50 & \textbf{81.09} & 1.71 & 6.90 \\
\bottomrule
\end{tabular}
\label{tab:tab2}
\end{table*}

\subsection{Implementation Details and Evaluation Metrics}

All experiments were implemented in PyTorch and executed on the NVIDIA 4080 GPU. Model training used the SGD optimizer for 17000 iterations, with an initial learning rate of 0.01 and a batch size of 4. Our model was designed to operate within a sliding window paradigm. For the LA dataset, inputs were randomly cropped sub-volumes of 112 × 112 × 80 voxels, extracted with a spatial stride of 18 × 18 × 4 voxels. For the Pancreas-CT and BraTS2019 dataset, a sampling window of 96 × 96 × 96 voxels was utilized, with a spatial stride of 16 × 16 × 16 voxels. We adopt three complementary evaluation metrics: the Dice Similarity Coefficient (DSC), the 95\% Hausdorff Distance (95HD), and the Average Surface Distance (ASD). DSC quantifies the volumetric overlap via intersection over union. ASD elucidates the average surface alignment precision by computing the mean of all point-to-surface distances; 95HD focuses on capturing robust local maximum boundary deviations by considering the 95th percentile of the Hausdorff Distance. Thus, DSC evaluates global shape similarity, ASD quantifies average boundary conformity, and 95HD is sensitive to clinically relevant maximum segmentation errors.

\subsection{Evaluation Protocol and Comparison Standards}
\label{sec:csd}

Evaluation protocols in SSL diverge into two paradigms. The first reports peak performance by selecting the optimal checkpoint based on evaluation metrics. Crucially, many existing methods achieve this by directly evaluating during training without a separate validation set. This practice introduces a significant bias, as the reported best results essentially reflect overfitting to the specific test distribution rather than true generalization. In contrast, the second paradigm evaluates performance at the final training checkpoint. To ensure a rigorous and fair assessment, our comparisons focus exclusively on methods evaluated at their final training checkpoints. We exclude methods that utilize the best iteration, as such protocols rely on information unavailable in real-world deployment and overestimate the model's practical capability.

\subsection{Comparison with Sate-of-the-art Methods}

\begin{figure*}[t]
\includegraphics[width=1.0\linewidth]{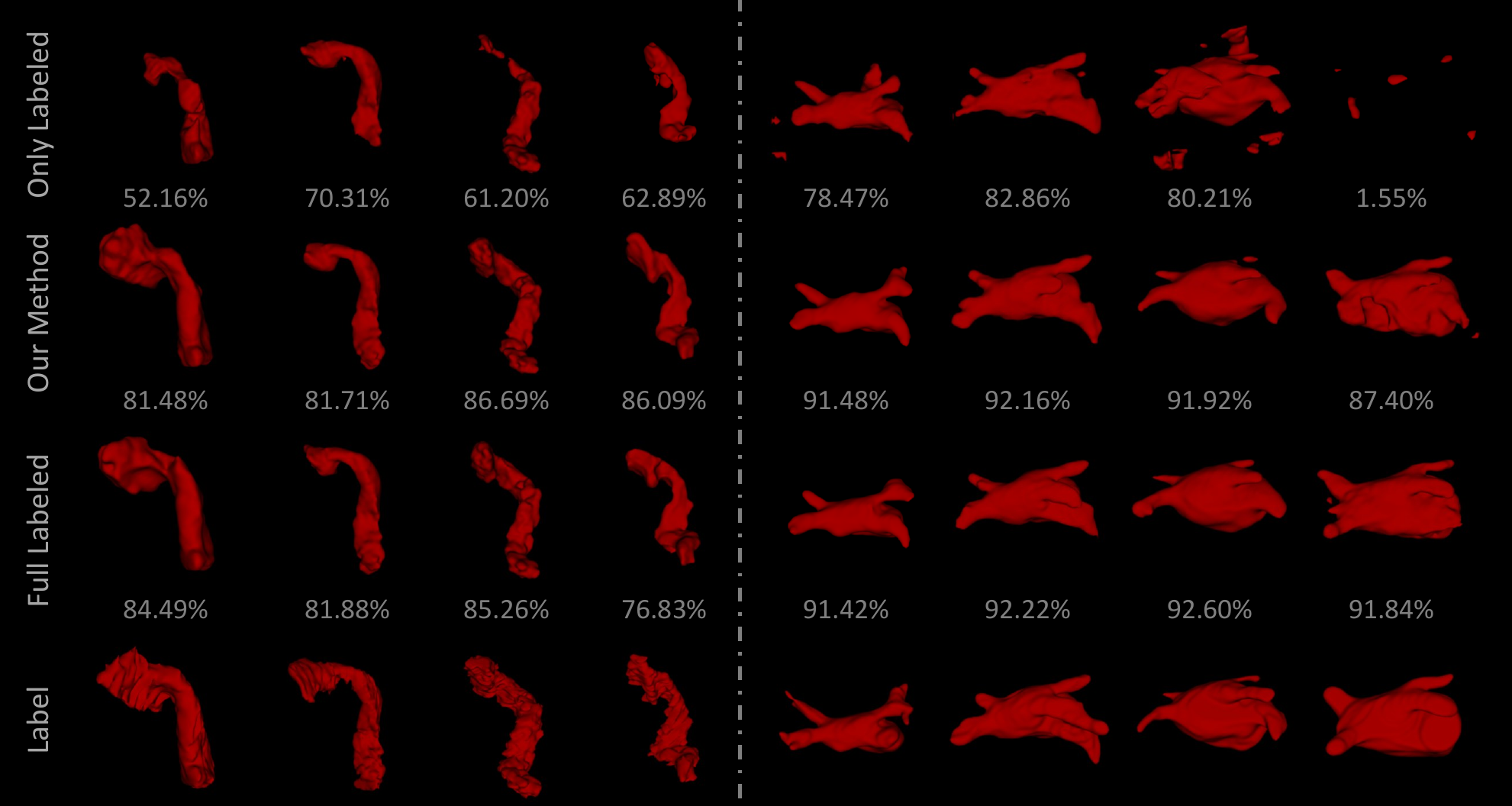}
\centering
\caption{3D Visual comparison of segmentation results with varying label proportions.}
\label{fig:fig6}
\end{figure*}

We conducted comparative evaluations against several state-of-the-art SSL methods in Table \ref{tab:tab1}, Table \ref{tab:tab2} and Table \ref{tab:tabbrats}, including DTC \cite{luo2021semi}, DUWM \cite{wang2020double}, AC-MT \cite{xu2023ambiguity}, MCF \cite{wang2023mcf}, CoraNet \cite{shi2021inconsistency}, TAC \cite{chen2022semi}, BaPC \cite{wang2024boundary}, UG-MCL \cite{zhang2023uncertainty}, AUSS \cite{adiga2024anatomically}, SCC \cite{liu2022contrastive}, UA-MT \cite{yu2019uncertainty}, URPC \cite{luo2022semi}, AUA \cite{wu2023exploring}, SASSNet \cite{li2020shape}, CPCL \cite{9741294}, MRPL \cite{su2024mutual} and CauSSL \cite{miao2023caussl}. To minimize confounding factors from varying experimental setups across studies, we directly adopted performance metrics reported in the original publications. As shown in Table~\ref{tab:tab1}, DBiSL achieves the best performance on LA (90.54\% Dice/1.80 ASD/6.05 95HD), consistently outperforming prior SSL baselines. On Pancreas-CT (Table~\ref{tab:tab2}), DBiSL attains the highest Dice score (81.09\%), while remaining competitive on boundary metrics. Importantly, Table~\ref{tab:tabbrats} further demonstrates strong generalization to brain MRI, where DBiSL achieves the best overall results on BraTS2019 (85.09\% Dice/1.89 ASD/8.12 95HD). In addition, Figure ~\ref{fig:fig6} provides 3D visual comparisons against fully supervised and semi-supervised baselines, qualitatively confirming that our differentiable bidirectional collaboration yields more accurate and structurally consistent boundaries. Overall, these results validate the effectiveness and robustness of our synergistic learning framework enabled by the proposed fully differentiable bidirectional transformer.

\begin{table*}[t]
\centering
\caption{Comparisons with other semi-supervised methods on BraTS2019 dataset.}
\begin{tabular}{lccccc}
\toprule
\multirow{2}{*}{Method} & \multicolumn{2}{c}{Scans used} & \multicolumn{3}{c}{Metrics} \\
\cmidrule(lr){2-3} \cmidrule(lr){4-6} 
& Labeled & Unlabeled & Dice (\%) $\uparrow$ & ASD (mm) $\downarrow$ & 95HD (mm) $\downarrow$ \\
\midrule
Supervised baseline & 25 & 0 & 76.29 & 6.63 & 22.98 \\
Supervised baseline & 250 & 0 & 85.13 & 2.14 & 8.39 \\
\midrule
UG-MCL \cite{zhang2023uncertainty} & 25 & 225 & 82.82 & 2.30 & 11.29 \\
CPCL \cite{9741294} & 25 & 225 & 83.36 & 1.99 & 11.74 \\
CauSSL \cite{miao2023caussl} & 25 & 225 & 83.54 & 1.98 & 12.53 \\
AC-MT \cite{xu2023ambiguity} & 25 & 225 & 83.77 & 1.93 & 11.37 \\
URPC  \cite{luo2022semi} & 25 & 225 & 84.16 & 2.63 & 11.01 \\
MRPL \cite{su2024mutual} & 25 & 225 & 84.29 & 2.55 & 9.57 \\
\midrule
DBiSL & 25 & 225 & \textbf{85.09} & \textbf{1.89} & \textbf{8.12} \\
\bottomrule
\end{tabular}
\label{tab:tabbrats}
\end{table*}

\subsection{Ablation Study on Key Components}

\begin{figure*}[t]
\includegraphics[width=1.0\linewidth]{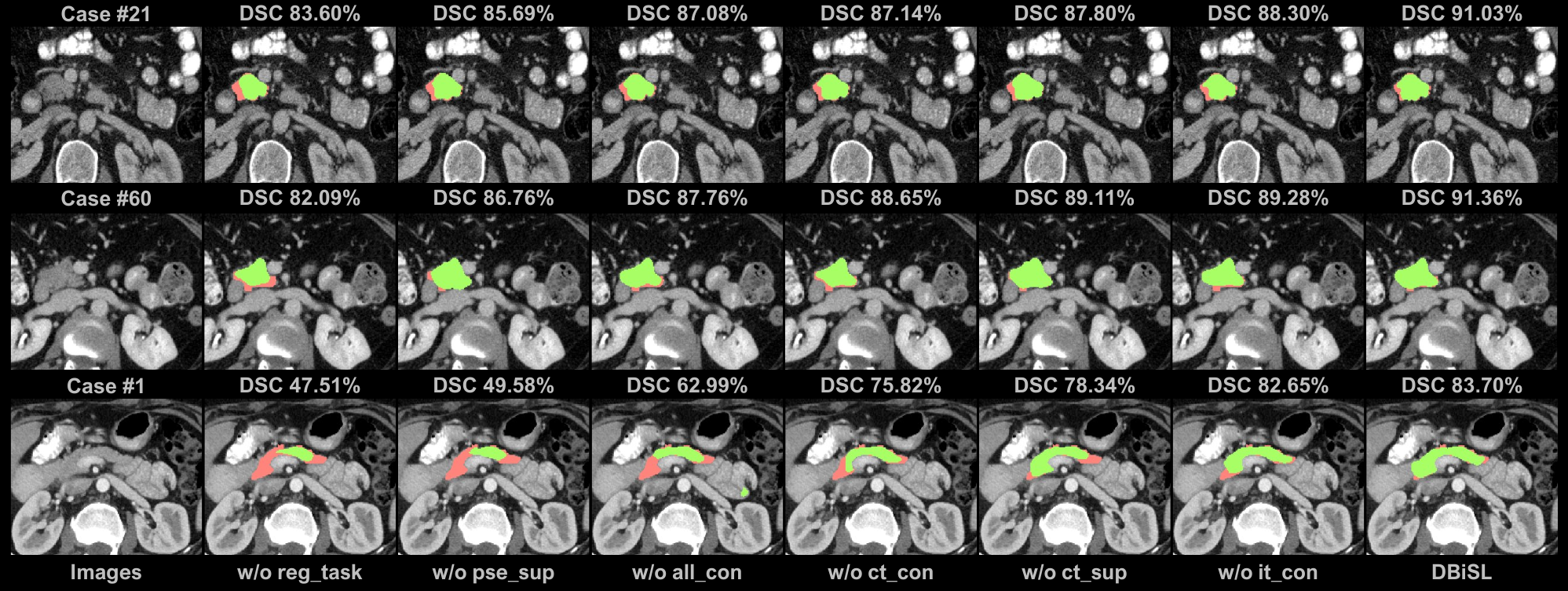}
\centering
\caption{Visual comparison of results from the ablation study. Green and red regions delineate the outputs and the ground truth, respectively.}
\label{fig:fig3}
\end{figure*}

To dissect the contribution of key components, we conducted the ablation study (Table \ref{tab:tab3}, Figure~\ref{fig:fig3}), evaluating the impact of: \textbf{reg\_task} (all distance regression tasks); \textbf{pse\_sup} (pseudo-supervised learning); \textbf{ct\_sup} (cross-task supervised learning); \textbf{all\_con} (all consistency learning); \textbf{ct\_con} (cross-task consistency); and \textbf{it\_con} (intra-task consistency). Ablation results reveal that the regression task most significantly enhances performance, strongly validating multi-task learning for SSL. Pseudo-supervision and consistency regularization, as the most popular techniques, also substantially improve performance. The cross-task supervision in the supervised learning is empirically validated to effectively enhance performance. These findings clearly demonstrate DBiSL's synergistic nature: removing any key component would degrade performance, highlighting the critical and synergistic value of modules.

% \subsection{Distance regression results}

\subsection{Ablation Study on Unified Processing}

A typical SSL approach treats labeled and unlabeled data disparately: labeled data for supervised learning, unlabeled data solely for pseudo-supervision. The rationale is that labeled data, possessing inherent label information, seemingly negates the need for pseudo-supervision. Nevertheless, such disjointed paradigms can engender a "shortcut" effect, resulting in a processing imbalance. Thus, we tested removing labeled data from pseudo-supervision, the results in Table \ref{tab:tab4} show quantifiable performance degradation. Thus, a unified processing approach facilitates a more holistic utilization of information from all available data, mitigating biases toward specific data types.

\subsection{Performance on the unlabeled training data}

To evaluate pseudo-label quality, we conducted a post-hoc analysis on the unlabeled training data (Table \ref{tab:tab5}). For the LA dataset, the model achieves considerable accuracy despite limited supervision, with high-confidence masks significantly improving precision by mitigating erroneous signals. Conversely, due to the inherent segmentation challenges associated with the pancreas dataset, performance on the pancreas dataset remains inferior to the LA results, underscoring the need for specialized segmentation strategies. Consequently, future research endeavors should prioritize the exploration and integration of customized strategies specifically tailored for pancreas segmentation tasks. Overall, the model’s robust performance on unlabeled data validates the effective transition from a semi-supervised to an approximate fully-supervised paradigm.

\begin{table}[t]
\centering
\caption{Ablation study on Pancreas-CT dataset.}
\begin{tabular}{cccc}
\toprule
\multirow{3.5}{*}{Method} & \multicolumn{3}{c}{Metrics} \\ 
\cmidrule{2-4} 
& \makecell{Dice\\(\%)} $\uparrow$ & \makecell{ASD\\(mm)} $\downarrow$ & \makecell{95HD\\(mm)} $\downarrow$ \\
\midrule
DBiSL & 81.09 & 1.71 & 6.90 \\
w/o reg\_task & 64.58 & 1.11 & 18.19 \\
w/o pse\_sup & 66.45 & 4.13 & 13.82 \\
w/o ct\_sup & 76.32 & 2.47 & 10.66 \\
w/o all\_con & 68.15 & 1.29 & 20.33 \\
w/o ct\_con & 71.57 & 1.25 & 14.43 \\
w/o it\_con & 76.83 & 1.24 & 9.37 \\
\bottomrule
\end{tabular}
\label{tab:tab3}
\end{table}

\begin{table}[!h]
\centering
\caption{Ablation study on unified processing.}
\begin{tabular}{cccc}
\toprule
\multirow{3.5}{*}{Method} & \multicolumn{3}{c}{Metrics} \\ 
\cmidrule{2-4} 
& \makecell{Dice\\(\%)} $\uparrow$ & \makecell{ASD\\(mm)} $\downarrow$ & \makecell{95HD\\(mm)} $\downarrow$ \\
\midrule
with labeled data & 90.54 & 1.80 & 6.05 \\
w/o labeled data & 89.40 & 1.81 & 6.78 \\
\bottomrule
\end{tabular}
\label{tab:tab4}
\end{table}

\begin{table}[!h]
\centering
\caption{Results for unlabeled training samples.}
\begin{tabular}{cccc}
\toprule
\multirow{3.5}{*}{Dataset} & \multicolumn{3}{c}{Metrics} \\ 
\cmidrule{2-4} 
& \makecell{Dice\\(\%)} $\uparrow$ & \makecell{Recall\\(\%)} $\uparrow$ & \makecell{Precision\\(\%)} $\uparrow$ \\
\midrule
LA & 90.11 & 88.60 & 92.26 \\
Pancreas\_CT & 75.57 & 81.81 & 72.03 \\
\bottomrule
\end{tabular}
\label{tab:tab5}
\end{table}

\begin{table}[t]
\centering
\small
\caption{Comparison of offline transformation method and our proposed online transformation method on LA dataset.}
\begin{tabular}{ccccc}
\toprule
Method & \makecell{DSC \\ (\%)} & \makecell{Time \\ (s/case)} & Diff. & \makecell{GPU} \\
\midrule
offline-original & 100.00 & 1.03 & No & No \\
online-original & 99.95 & 0.05 & Yes & Yes \\
offline-resample & 91.17 & 0.10 & No & No \\
online-resample & 97.53 & 0.01 & Yes & Yes \\
\bottomrule
\end{tabular}
\label{tab:tab6}
\end{table}

\begin{figure}[t]
\centering
\includegraphics[width=1.0\columnwidth]{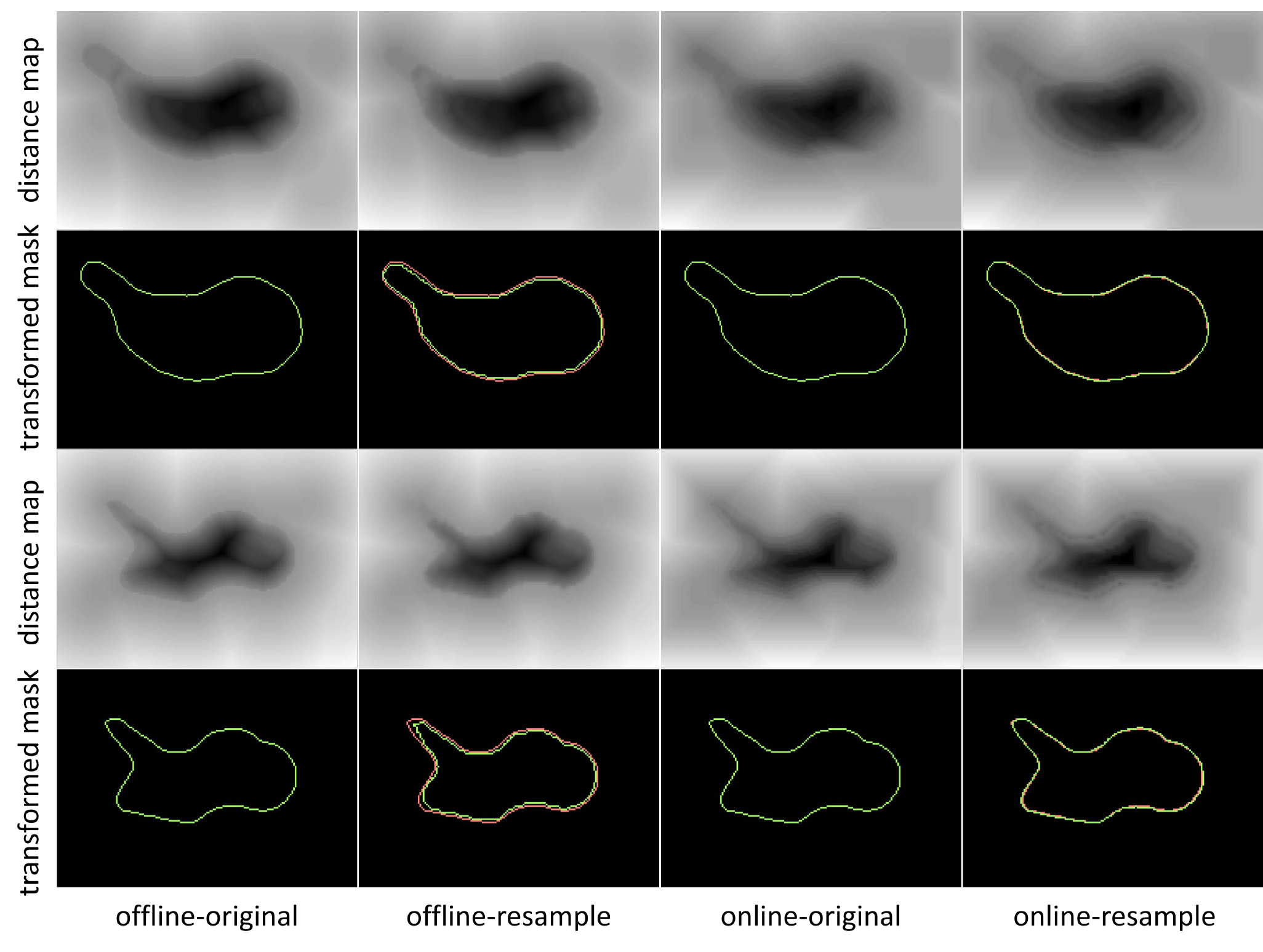}
\caption{Visualization of different distance transform methods. Green and red contours denote transformed and label contours, respectively.}
\label{fig:fig4}
\end{figure}

\subsection{Performance under Various Labeling Ratios}
To test the model's stability, we evaluated DBiSL on the LA dataset using 5\%, 10\%, 20\%, and 50\% labeled data. As shown in Table \ref{tab:tabrat}, we compared our results against several SOTA methods. For a fair comparison, the performance metrics for these methods were taken directly from their original reports. The results show that DBiSL consistently leads other SOTA methods across all ratios. Notably, even with only 5\% labels, our method achieves a Dice score of 84.62\%, outperforming the baseline by a wide margin of 22.39\%. This suggests that our bidirectional synergistic learning is particularly effective when labeled data is extremely scarce. While the performance gap narrows at 50\% supervision, DBiSL still maintains an edge, proving its robustness across varying levels of data availability.

\begin{table}[t]
\centering
\caption{Results with various labeling ratios.}
\begin{tabular}{ccccc}
\toprule
\multirow{2.5}{*}{Method} & \multicolumn{4}{c}{Labeling Ratios} \\ 
\cmidrule{2-5} 
& 5\% & 10\% & 20\% & 50\% \\
\midrule
Baseline & 62.23 & 79.44 & 82.36 & 91.50 \\
TAC & / & 84.73 &	87.75 &	/ \\
AUSS & / & 86.58 &	88.6 &	/ \\
MCF & / & / &	88.71 &	/ \\
UAMT & / & 84.25 &	88.88 &	/ \\
DUWM & / & 85.91 &	89.65 &	/ \\
DTC & / & 87.51 &	89.42 &	/ \\
SASSNet & / & 86.81 &	89.27 &	/ \\
BaPC & / & 88.55 &	89.71 &	/ \\
SCC & / & 86.51 &	89.81 &	/ \\
DBiSL & 84.62 & 88.78 & 90.54 & 91.84 \\
\bottomrule
\end{tabular}
\label{tab:tabrat}
\end{table}

\subsection{Precision and Computational Efficiency of Different Distance Transformation Methods}

We validated our differentiable task transformer against a widely used offline method. By leveraging convolutional operations, our transformer is GPU‑accelerated, offering substantial computational advantages over CPU‑bound offline approaches. To alleviate computational bottlenecks, we downsample the binary mask, apply the distance transformer, and then upsample the output to the original resolution (denoted as resample in Table \ref{tab:tab6}). It should be noted that our distance approximation, implemented with customized convolutional kernels, inherently introduces some precision loss. We quantified this loss by inversely transforming distance maps to segmentation masks and evaluating the DSC against ground‑truth labels. As shown in Table \ref{tab:tab6} and Figure~\ref{fig:fig4}, the precision loss remains within acceptable margins. Moreover, because the same differentiable transformation is used in both the regression label generation and model inference, the approximation error originates from a consistent source, thereby reducing its potential adverse impact. Thus, the inherent precision loss is not expected to critically affect model performance.

\subsection{Impact of Downsampling Rate}

Maintaining the propagation of gradient information typically incurs significant GPU memory overhead. We addressed this by performing downsampling on the input prior to online distance map generation. Although we verified in the previous section that this downsampling operation has a limited impact on the intrinsic accuracy of the distance map itself, its effect on the model's overall performance warrants further investigation. To this end, we explored different downsampling scale factors to assess the specific influence of this operation on the model's segmentation performance. Notably, using the original full resolution for differentiable distance transform during training results in GPU memory overflow on an RTX 4080. Consequently, we set the upper bound for the downsampling scale factor to 0.5. Experimental results, presented in Table \ref{tab:tab8}, are consistent with intuition: model performance is optimal when the downsampling scale factor is set to 0.5. This indicates that a lower degree of downsampling (i.e., a higher scale factor) results in less information loss introduced by the distance transform, thereby yielding higher model performance. Furthermore, it is noteworthy that the model exhibits relatively low sensitivity to this downsampling scale factor hyperparameter; the model's overall performance remains at a high level.

\begin{table}[t]
\caption{Segmentation performance with different downsampling rate on LA dataset.}
\centering
\begin{tabular}{cccc}
\toprule
\multirow{3}{*}{Down Rate} & \multicolumn{3}{c}{Metrics} \\
\cmidrule{2-4}
 &  \makecell{Dice \\ (\%)} $\uparrow$ & \makecell{ASD \\ (mm)} $\downarrow$ & \makecell{95HD \\ (mm)} $\downarrow$ \\
\midrule
0.5 &  90.54 & 1.80 & 6.05 \\
0.3 & 90.40 & 1.59 & 6.69 \\
0.1 & 90.01 & 1.58 & 6.83 \\
\bottomrule
\end{tabular}
\label{tab:tab8}
\end{table}

\begin{table}[t]
\caption{Performance with Pre-generated and On-the-fly Distance Maps on Pancreas-CT Dataset.}
\centering
\begin{tabular}{cccc}
\toprule
\multirow{3}{*}{Method} & \multicolumn{3}{c}{Metrics} \\
\cmidrule{2-4}
&  \makecell{Dice \\ (\%)} $\uparrow$ & \makecell{ASD \\ (mm)} $\downarrow$ & \makecell{95HD \\ (mm)} $\downarrow$ \\
\midrule
On-the-fly & 81.09 & 1.71 & 6.90 \\
Pre-generated  & 74.48 & 1.42 & 8.52 \\
\bottomrule
\end{tabular}
\label{tab:tab7}
\end{table}

\begin{table}[t]
\centering
\caption{Computational costs of different methods on BraTS2019 dataset. G-Mem: GPU memory usage during training.}
\begin{tabular}{lcccc}
\toprule
Method & \makecell{Params \\ (M)} & \makecell{Train \\ (s/iter)} & \makecell{Test \\ (s/case)}  & \makecell{G-Mem \\ (GB)} \\
\midrule
DTC & 9.45 & 0.645 & 1.02 & 4.62 \\
CauSSL & 23.60 & 0.442 & 1.75 & 8.43 \\
UA-MT & 9.45 & 0.367 & 1.08 & 5.33 \\
MRPL & 12.35 & 0.334 & 1.67 & 10.40 \\
DBiSL & 12.35 & 0.492 & 1.66 & 9.49 \\
\bottomrule
\end{tabular}
\label{tab:tabcom}
\end{table}

\subsection{The Impact of Pre-Generation and On-the-fly Generation Schemes}

\begin{figure}[t]
\centering
\includegraphics[width=1.0\columnwidth]{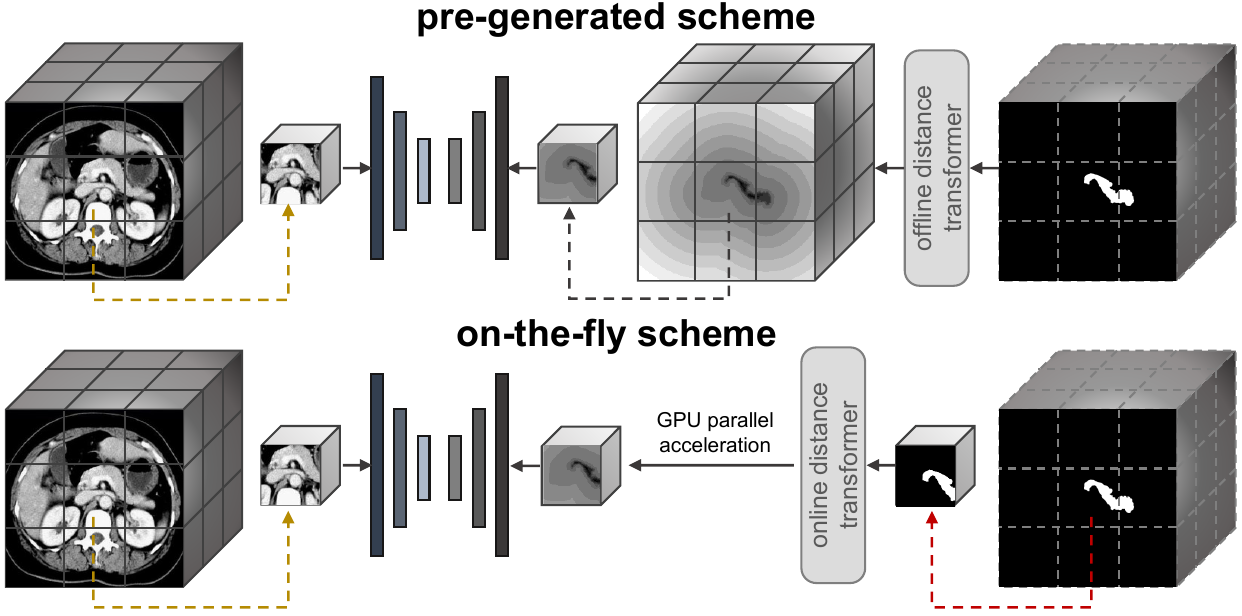}
\caption{Workflow of pre-generated and on-the-fly schemes. The former selects corresponding patches from pre-generated full-resolution distance maps as labels for the distance regression task, while the latter generates distance labels based on the respective patches.}
\label{fig:fig7}
\end{figure}

Existing multi-task schemes, which often integrate distance regression, are generally inefficient due to their reliance on offline distance map generation, failing to fully utilize modern GPU parallel computing power. While this pre-generation strategy is feasible for 2D segmentation (where targets are computed based on the full image regardless of whether they are generated offline or online), it introduces a critical challenge in 3D patch-based segmentation. Due to GPU memory constraints, mainstream 3D methods use a sliding window strategy, processing only a local input patch. This leads to an inherent information asymmetry: offline targets are computed from the entire 3D volume, whereas online predictions are made solely based on the local patch information. This requires the model to infer global distance values from local context, significantly complicating the distance regression task. Our GPU-accelerated differentiable transformer provides an efficient generation paradigm that addresses this limitation. By enabling online, GPU-accelerated computation, it effectively alleviates the information asymmetry issue inherent in existing approaches (Figure~\ref{fig:fig7}). Comparative results (Table \ref{tab:tab7}) confirm the superiority of our proposed scheme.

\subsection{Computational Efficiency of Different Models.}
To assess the practicality of DBiSL, we report its computational footprint in Table~\ref{tab:tabcom} and benchmark it against strong SSL competitors. Overall, DBiSL keeps a compact architecture (12.35M parameters), comparable to MRPL and notably lighter than CauSSL (23.60M). The differentiable bidirectional coupling introduces additional computation during training, resulting in 0.492 s/iter; nevertheless, this cost remains below DTC (0.645 s/iter) and is within the same practical regime as other modern SSL baselines. Importantly, the proposed design does not inflate inference latency: DBiSL runs at 1.66 s/case, essentially matching MRPL (1.67 s/case) and slightly improving over CauSSL (1.75 s/case). Regarding memory, DBiSL requires 9.49 GB during training—higher than lightweight consistency baselines such as UA-MT (5.33 GB), yet lower than MRPL (10.40 GB)—and thus remains compatible with commonly available GPUs for 3D segmentation. These results suggest that the additional differentiability-enabled collaboration yields improved accuracy while preserving a practically deployable computational profile.

\begin{table}[t]
\centering
\caption{Performance with four backbones on LA dataset.}
\begin{tabular}{lcccc}
\toprule
Backbone & \makecell{DSC \\ (\%)} & \makecell{ASD \\ (mm)} & \makecell{95HD \\ (mm)} & \makecell{Params \\ (M)} \\
\midrule
V-Net & 90.54 & 1.80 & 6.05 & 12.35 \\
nnU-Net & 90.09 & 2.45 & 11.69 & 48.36 \\
DiffUNet & 90.13 & 2.63 & 11.56 & 80.18 \\
SegMamba & 88.88 & 3.15 & 11.61 & 86.87 \\
\bottomrule
\end{tabular}
\label{tab:backbone_perf}
\vspace{1mm}
{\footnotesize\raggedright
We reduce the patch size to $64{\times}64{\times}64$ for SegMamba and $96{\times}96{\times}96$ for nnU-Net and DiffUNet to avoid overflow.
\par}
\end{table}

\subsection{Backbone Generality.}
We further instantiate DBiSL with nnU-Net \cite{isensee2021nnu}, DiffUNet \cite{xing2025diff}, and SegMamba \cite{xing2025segmamba} on the LA dataset under the same labeled/unlabeled split. As shown in Table~\ref{tab:backbone_perf}, replacing V-Net with nnU-Net or DiffUNet yields comparable Dice scores (90.09\%/90.13\% vs.\ 90.54\%), supporting the plug-and-play nature of our SSL paradigm. We note that, due to GPU memory constraints, we reduce the patch size to $96{\times}96{\times}96$ for nnU-Net and DiffUNet and to $64{\times}64{\times}64$ for SegMamba; the smaller 3D context may weaken global anatomical continuity modeling and thus negatively impact boundary-sensitive metrics, which likely contributes to the relatively lower performance of SegMamba in our current setting. Overall, these results indicate that the gains of DBiSL are not specific to a particular network design, but stem from the proposed learning scheme. At the same time, they suggest that extracting the best performance from large-capacity backbones in extremely low-label settings can be sensitive to resource-dependent choices (e.g., patch size) and backbone-specific training recipes.

\section{Discussion}

\subsection{Difference from Unidirectional Segmentation-regression Synergistic Methods}
\label{sec:dis}

This work is closely related to DTC~\cite{luo2021semi} in that both exploit the complementarity between segmentation and distance regression under SSL and are evaluated on overlapping benchmarks. However, DTC essentially yields a unidirectional synergy, where the auxiliary regression branch mainly provides guidance to improve segmentation. Such one-way interaction prevents segmentation-derived structural cues (e.g., explicit boundary/topology information) from being fed back to refine distance regression during training, thereby limiting mutual task reinforcement. In contrast, we introduce a fully differentiable bidirectional task transformer that enables end-to-end, reciprocal information exchange between the two tasks (regression$\rightarrow$segmentation and segmentation$\rightarrow$regression). Targeted ablations (Table~\ref{tab:taboneway}) further verify the necessity of bidirectionality: DBiSL achieves 81.09\% DSC with 6.90~mm 95HD, while the fully differentiable unidirectional variants drop markedly to 68.79\%/17.65~mm (DBiSL-S2R) and 71.59\%/16.77~mm (DBiSL-R2S), indicating that one-way guidance is insufficient to resolve the mutual dependency and latent conflicts between the two tasks under scarce supervision.

\begin{figure*}[t]
\centering
\includegraphics[width=1.0\linewidth]{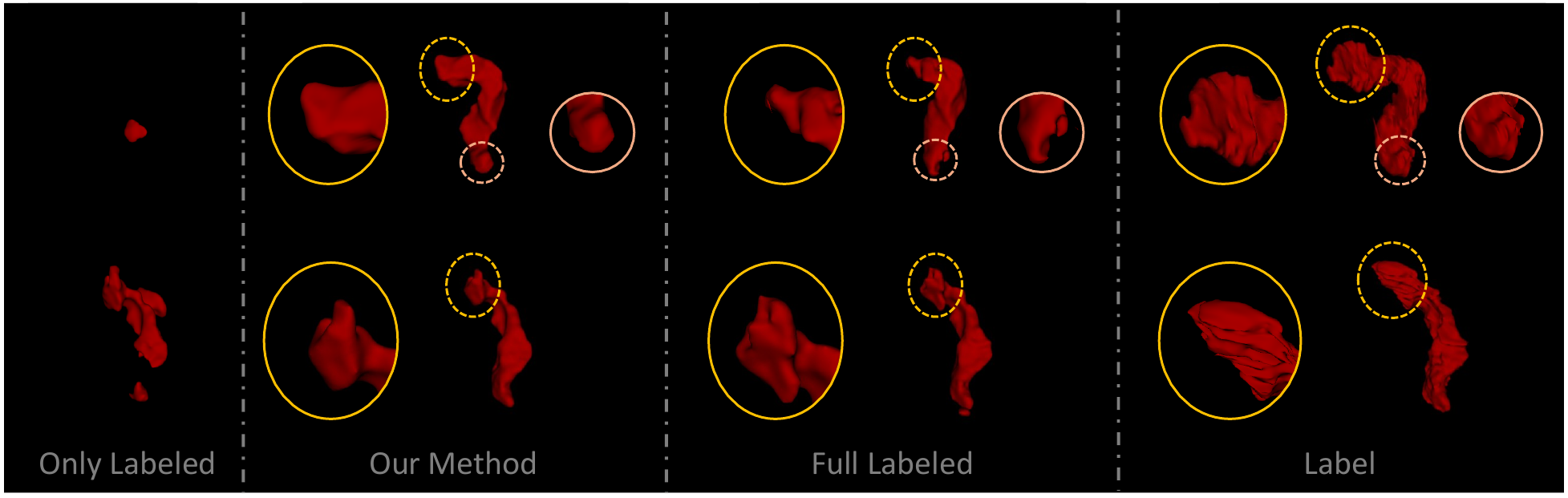}
\caption{3D visualization of two cases demonstrating poor segmentation performance.}
\label{fig:fig5}
\end{figure*}

\begin{table}[t]
\caption{Performance of unidirectional and bidirectional synergistic methods.}
\centering
\begin{tabular}{cccc}
\toprule
\multirow{3}{*}{Method} & \multicolumn{3}{c}{Metrics} \\
\cmidrule{2-4}
& \makecell{Dice \\ (\%)} $\uparrow$ & \makecell{ASD \\ (mm)} $\downarrow$ & \makecell{95HD \\ (mm)} $\downarrow$ \\
\midrule
DBiSL &  81.09 & 1.71 & 6.90 \\
DBiSL-S2R  & 68.79 & 1.16 & 17.65 \\
DBiSL-R2S  & 71.59 & 1.12 & 16.77 \\
\bottomrule
\end{tabular}
\label{tab:taboneway}
\end{table}

\subsection{Difference from Other Multi-task Learning Methods}

This work is distinguished from existing dual-task driven methods in three key aspects. First, for the synergy mechanism, our DBiSL achieves bidirectional information interaction, as elaborated in Sec.\ref{sec:dis}. Second, for the framework construction, DBiSL presents a comprehensive and self-consistent framework. It integrates essential SSL components—pseudo-supervision, confidence estimation, consistency regularization, and supervised learning—overcoming limitations of isolated techniques by unifying them under a coherent objective. Third, for the training paradigms, we leverage GPU parallelism for real-time patch-based distance map generation within each mini-batch. This contrasts with offline pre-computation in prior methods, which, while reducing CPU load in training, forces models to infer global context from local patch—potentially hindering training.

\subsection{Analysis of Failure Cases}

Figure \ref{fig:fig5} illustrates the segmentation results of our proposed method, the baseline (Only Labeled), and the theoretical upper bound (Full Labeled) on two challenging samples. As observed, both our method and the Full Labeled approach achieve relatively accurate segmentation of the pancreatic body. However, their segmentation performance significantly degrades when it comes to the head and tail regions of the pancreas. This phenomenon is primarily attributed to the high variability of the pancreatic anatomical structure and its indistinct boundaries with surrounding tissues \cite{10982257}. Specifically, the pancreatic body, being the central part of the organ, benefits from ample anatomical context information. In contrast, the pancreatic head and tail, located at the periphery, suffer from a scarcity of reliable 3D contextual cues. This poses a challenge for our distance regression task, which struggles to maintain precision in these regions due to the presence of blurry boundaries and limited spatial information.

\subsection{Limitations and Future Works}
Despite the strong performance of DBiSL, several limitations warrant future study. Specifically, our experiments focus on Whole Tumor segmentation on BraTS 2019 to follow the canonical semi-supervised benchmarking protocol and enable fair comparisons with prior work. This evaluation does not cover clinically important multi-class targets (e.g., Tumor Core and Enhancing Tumor). Technically, DBiSL readily generalizes to multi-class segmentation by increasing the output channels of the segmentation/regression heads and applying the task-transformation modules in a channel-wise manner, while keeping the bidirectional interaction and synergistic losses unchanged. Future work will extend DBiSL to more realistic settings, including multi targets, incomplete annotations and modalities, to improve robustness and clinical utility under heterogeneous data conditions.

\section{Conclusion}

In this work, we introduced DBiSL, a novel differentiable bidirectional synergistic learning framework for semi-supervised 3D medical image segmentation. It overcomes limitations of unidirectional dual-task methods via a fully differentiable bidirectional transformer, enabling genuine online inter-task bidirectional synergy. Unifying supervised learning, consistency regularization, uncertainty estimation, and pseudo-supervision within a coherent architecture, DBiSL synergistically boosts segmentation performance under severe label scarcity, achieving state-of-the-art results on two benchmarks. Beyond performance, DBiSL provides key insights into unified SSL framework design via dual-task collaboration, establishing a robust bidirectional interaction paradigm for future SSL and broader multi-task vision applications.

%%%%%%%%%%%%%%%%%%%%%%%%%%%%%%%%%%%%%%%%%%%%%%%%%%%%%%%%%%%%%%%%%%%%%%%%
\section*{Acknowledgements}
This work was supported by the National Natural Science Foundation of China under Grant 62401481, Natural Science Foundation of Sichuan Province under Grant 2025ZNSFSC1450, China Postdoctoral Science Foundation under Grant 2024M752683.

%%%%%%%%%%%%%%%%%%%%%%%%%%%%%%%%%%%%%%%%%%%%%%%%%%%%%%%%%%%%%%%%%%%%%%%%
\bibliography{biblio}

@article{li2022dual,
  title={A dual meta-learning framework based on idle data for enhancing segmentation of pancreatic cancer},
  author={Li, Jun and Qi, Liang and Chen, Qingzhong and Zhang, Yu-Dong and Qian, Xiaohua},
  journal={Medical Image Analysis},
  volume={78},
  pages={102342},
  year={2022},
  publisher={Elsevier}
}

@article{su2024mutual,
  title={Mutual learning with reliable pseudo label for semi-supervised medical image segmentation},
  author={Su, Jiawei and Luo, Zhiming and Lian, Sheng and Lin, Dazhen and Li, Shaozi},
  journal={Medical Image Analysis},
  volume={94},
  pages={103111},
  year={2024},
  publisher={Elsevier}
}

@ARTICLE{10486949,
  author={Xu, Chenchu and Zhang, Tong and Zhang, Dong and Zhang, Dingwen and Han, Junwei},
  journal={IEEE Transactions on Medical Imaging}, 
  title={Deep Generative Adversarial Reinforcement Learning for Semi-Supervised Segmentation of Low-Contrast and Small Objects in Medical Images}, 
  year={2024},
  volume={43},
  number={9},
  pages={3072-3084},
  doi={10.1109/TMI.2024.3383716}}

@ARTICLE{10982257,
  author={Li, Jun and Zhang, Yijue and Shi, Haibo and Li, Minhong and Li, Qiwei and Qian, Xiaohua},
  journal={IEEE Transactions on Medical Imaging}, 
  title={A Dual-Task Synergy-Driven Generalization Framework for Pancreatic Cancer Segmentation in CT Scans}, 
  year={2025},
  volume={44},
  number={9},
  pages={3783-3794},
  doi={10.1109/TMI.2025.3566376}}

@inproceedings{wang2024semi,
  title={Semi-supervised Medical Image Segmentation with Strong/Weak Task-Aware Consistency},
  author={Wang, Hua and Qiu, Linwei and Li, Yiming and Hu, Jingfei and Zhang, Jicong},
  booktitle={Chinese Conference on Pattern Recognition and Computer Vision (PRCV)},
  pages={17--31},
  year={2024},
  organization={Springer}
}

@article{ZHU2024102383,
title = {Hybrid dual mean-teacher network with double-uncertainty guidance for semi-supervised segmentation of magnetic resonance images},
journal = {Computerized Medical Imaging and Graphics},
volume = {115},
pages = {102383},
year = {2024},
issn = {0895-6111},
doi = {https://doi.org/10.1016/j.compmedimag.2024.102383},
url = {https://www.sciencedirect.com/science/article/pii/S0895611124000600},
author = {Jiayi Zhu and Bart Bolsterlee and Brian V.Y. Chow and Yang Song and Erik Meijering}
}

@article{HAN2024123052,
title = {Deep semi-supervised learning for medical image segmentation: A review},
journal = {Expert Systems with Applications},
volume = {245},
pages = {123052},
year = {2024},
issn = {0957-4174},
doi = {https://doi.org/10.1016/j.eswa.2023.123052},
url = {https://www.sciencedirect.com/science/article/pii/S0957417423035546},
author = {Kai Han and Victor S. Sheng and Yuqing Song and Yi Liu and Chengjian Qiu and Siqi Ma and Zhe Liu}
}

@article{WU2025103547,
title = {Medical SAM adapter: Adapting segment anything model for medical image segmentation},
journal = {Medical Image Analysis},
volume = {102},
pages = {103547},
year = {2025},
issn = {1361-8415},
doi = {https://doi.org/10.1016/j.media.2025.103547},
url = {https://www.sciencedirect.com/science/article/pii/S1361841525000945},
author = {Junde Wu and Ziyue Wang and Mingxuan Hong and Wei Ji and Huazhu Fu and Yanwu Xu and Min Xu and Yueming Jin}
}

@ARTICLE{10771804,
  author={Ran, Lingyan and Li, Yali and Liang, Guoqiang and Zhang, Yanning},
  journal={IEEE Transactions on Circuits and Systems for Video Technology}, 
  title={Pseudo Labeling Methods for Semi-Supervised Semantic Segmentation: A Review and Future Perspectives}, 
  year={2024},
  volume={},
  number={},
  pages={1-1},
  doi={10.1109/TCSVT.2024.3508768}}

@article{yang2025unimatch,
  title={Unimatch v2: Pushing the limit of semi-supervised semantic segmentation},
  author={Yang, Lihe and Zhao, Zhen and Zhao, Hengshuang},
  journal={IEEE Transactions on Pattern Analysis and Machine Intelligence},
  year={2025},
  publisher={IEEE}
}

@inproceedings{yu2019uncertainty,
  title={Uncertainty-aware self-ensembling model for semi-supervised 3D left atrium segmentation},
  author={Yu, Lequan and Wang, Shujun and Li, Xiaomeng and Fu, Chi-Wing and Heng, Pheng-Ann},
  booktitle={Medical image computing and computer assisted intervention--MICCAI 2019: 22nd international conference, Shenzhen, China, October 13--17, 2019, proceedings, part II 22},
  pages={605--613},
  year={2019},
  organization={Springer}
}

@article{XIONG2021101832,
title = {A global benchmark of algorithms for segmenting the left atrium from late gadolinium-enhanced cardiac magnetic resonance imaging},
journal = {Medical Image Analysis},
volume = {67},
pages = {101832},
year = {2021},
issn = {1361-8415},
doi = {https://doi.org/10.1016/j.media.2020.101832},
url = {https://www.sciencedirect.com/science/article/pii/S1361841520301961},
author = {Zhaohan Xiong and Qing Xia and Zhiqiang Hu and Ning Huang and Cheng Bian and Yefeng Zheng and Sulaiman Vesal and Nishant Ravikumar and Andreas Maier and Xin Yang and Pheng-Ann Heng and Dong Ni and Caizi Li and Qianqian Tong and Weixin Si and Elodie Puybareau and Younes Khoudli and Thierry Géraud and Chen Chen and Wenjia Bai and Daniel Rueckert and Lingchao Xu and Xiahai Zhuang and Xinzhe Luo and Shuman Jia and Maxime Sermesant and Yashu Liu and Kuanquan Wang and Davide Borra and Alessandro Masci and Cristiana Corsi and Coen {de Vente} and Mitko Veta and Rashed Karim and Chandrakanth Jayachandran Preetha and Sandy Engelhardt and Menyun Qiao and Yuanyuan Wang and Qian Tao and Marta Nuñez-Garcia and Oscar Camara and Nicolo Savioli and Pablo Lamata and Jichao Zhao}
}

@article{xing2025segmamba,
  author={Xing, Zhaohu and Ye, Tian and Yang, Yijun and Cai, Du and Gai, Baowen and Wu, Xiao-Jian and Gao, Feng and Zhu, Lei},
  journal={IEEE Transactions on Medical Imaging}, 
  title={SegMamba-V2: Long-Range Sequential Modeling Mamba for General 3-D Medical Image Segmentation}, 
  year={2026},
  volume={45},
  number={1},
  pages={4-15},
  doi={10.1109/TMI.2025.3589797}}

@article{xing2025diff,
  title={Diff-UNet: A diffusion embedded network for robust 3D medical image segmentation},
  author={Xing, Zhaohu and Wan, Liang and Fu, Huazhu and Yang, Guang and Yang, Yijun and Yu, Lequan and Lei, Baiying and Zhu, Lei},
  journal={Medical Image Analysis},
  pages={103654},
  year={2025},
  publisher={Elsevier}
}

@article{isensee2021nnu,
  title={nnU-Net: a self-configuring method for deep learning-based biomedical image segmentation},
  author={Isensee, Fabian and Jaeger, Paul F and Kohl, Simon AA and Petersen, Jens and Maier-Hein, Klaus H},
  journal={Nature methods},
  volume={18},
  number={2},
  pages={203--211},
  year={2021},
  publisher={Nature Publishing Group}
}

@inproceedings{milletari2016v,
  title={V-net: Fully convolutional neural networks for volumetric medical image segmentation},
  author={Milletari, Fausto and Navab, Nassir and Ahmadi, Seyed-Ahmad},
  booktitle={2016 fourth international conference on 3D vision (3DV)},
  pages={565--571},
  year={2016},
  organization={IEEE}
}

@article{menze2014multimodal,
  title={The multimodal brain tumor image segmentation benchmark (BRATS)},
  author={Menze, Bjoern H and Jakab, Andras and Bauer, Stefan and Kalpathy-Cramer, Jayashree and Farahani, Keyvan and Kirby, Justin and Burren, Yuliya and Porz, Nicole and Slotboom, Johannes and Wiest, Roland and others},
  journal={IEEE transactions on medical imaging},
  volume={34},
  number={10},
  pages={1993--2024},
  year={2014},
  publisher={IEEE}
}

@inproceedings{roth2015deeporgan,
  title={Deeporgan: Multi-level deep convolutional networks for automated pancreas segmentation},
  author={Roth, Holger R and Lu, Le and Farag, Amal and Shin, Hoo-Chang and Liu, Jiamin and Turkbey, Evrim B and Summers, Ronald M},
  booktitle={International conference on medical image computing and computer-assisted intervention},
  pages={556--564},
  year={2015},
  organization={Springer}
}

@inproceedings{luo2021semi,
  title={Semi-supervised medical image segmentation through dual-task consistency},
  author={Luo, Xiangde and Chen, Jieneng and Song, Tao and Wang, Guotai},
  booktitle={Proceedings of the AAAI conference on artificial intelligence},
  volume={35},
  number={10},
  pages={8801--8809},
  year={2021}
}

@ARTICLE{9741294,
  author={Xu, Zhe and Wang, Yixin and Lu, Donghuan and Yu, Lequan and Yan, Jiangpeng and Luo, Jie and Ma, Kai and Zheng, Yefeng and Tong, Raymond Kai-yu},
  journal={IEEE Journal of Biomedical and Health Informatics}, 
  title={All-Around Real Label Supervision: Cyclic Prototype Consistency Learning for Semi-Supervised Medical Image Segmentation}, 
  year={2022},
  volume={26},
  number={7},
  pages={3174-3184},
  doi={10.1109/JBHI.2022.3162043}}

@inproceedings{wang2020deep,
  title={Deep distance transform for tubular structure segmentation in ct scans},
  author={Wang, Yan and Wei, Xu and Liu, Fengze and Chen, Jieneng and Zhou, Yuyin and Shen, Wei and Fishman, Elliot K and Yuille, Alan L},
  booktitle={Proceedings of the IEEE/CVF Conference on Computer Vision and Pattern Recognition},
  pages={3833--3842},
  year={2020}
}

@inproceedings{riba2020kornia,
  title={Kornia: an open source differentiable computer vision library for pytorch},
  author={Riba, Edgar and Mishkin, Dmytro and Ponsa, Daniel and Rublee, Ethan and Bradski, Gary},
  booktitle={Proceedings of the IEEE/CVF Winter Conference on Applications of Computer Vision},
  pages={3674--3683},
  year={2020}
}

@article{hu2020automatic,
  title={Automatic pancreas segmentation in CT images with distance-based saliency-aware DenseASPP network},
  author={Hu, Peijun and Li, Xiang and Tian, Yu and Tang, Tianyu and Zhou, Tianshu and Bai, Xueli and Zhu, Shiqiang and Liang, Tingbo and Li, Jingsong},
  journal={IEEE journal of biomedical and health informatics},
  volume={25},
  number={5},
  pages={1601--1611},
  year={2020},
  publisher={IEEE}
}

@article{xu2023ambiguity,
  title={Ambiguity-selective consistency regularization for mean-teacher semi-supervised medical image segmentation},
  author={Xu, Zhe and Wang, Yixin and Lu, Donghuan and Luo, Xiangde and Yan, Jiangpeng and Zheng, Yefeng and Tong, Raymond Kai-yu},
  journal={Medical Image Analysis},
  volume={88},
  pages={102880},
  year={2023},
  publisher={Elsevier}
}

@article{chen2022semi,
  title={Semi-supervised unpaired medical image segmentation through task-affinity consistency},
  author={Chen, Jingkun and Zhang, Jianguo and Debattista, Kurt and Han, Jungong},
  journal={IEEE Transactions on Medical Imaging},
  volume={42},
  number={3},
  pages={594--605},
  year={2022},
  publisher={IEEE}
}

@inproceedings{wang2020double,
  title={Double-uncertainty weighted method for semi-supervised learning},
  author={Wang, Yixin and Zhang, Yao and Tian, Jiang and Zhong, Cheng and Shi, Zhongchao and Zhang, Yang and He, Zhiqiang},
  booktitle={Medical Image Computing and Computer Assisted Intervention--MICCAI 2020: 23rd International Conference, Lima, Peru, October 4--8, 2020, Proceedings, Part I 23},
  pages={542--551},
  year={2020},
  organization={Springer}
}

@article{zhang2023uncertainty,
  title={Uncertainty-guided mutual consistency learning for semi-supervised medical image segmentation},
  author={Zhang, Yichi and Jiao, Rushi and Liao, Qingcheng and Li, Dongyang and Zhang, Jicong},
  journal={Artificial Intelligence in Medicine},
  volume={138},
  pages={102476},
  year={2023},
  publisher={Elsevier}
}

@article{adiga2024anatomically,
  title={Anatomically-aware uncertainty for semi-supervised image segmentation},
  author={Adiga, Sukesh and Dolz, Jose and Lombaert, Herve},
  journal={Medical Image Analysis},
  volume={91},
  pages={103011},
  year={2024},
  publisher={Elsevier}
}

@article{liu2022contrastive,
  title={A contrastive consistency semi-supervised left atrium segmentation model},
  author={Liu, Yashu and Wang, Wei and Luo, Gongning and Wang, Kuanquan and Li, Shuo},
  journal={Computerized Medical Imaging and Graphics},
  volume={99},
  pages={102092},
  year={2022},
  publisher={Elsevier}
}

@article{li2023generalizable,
  title={Generalizable pancreas segmentation via a dual self-supervised learning framework},
  author={Li, Jun and Zhu, Hongzhang and Chen, Tao and Qian, Xiaohua},
  journal={IEEE Journal of Biomedical and Health Informatics},
  volume={27},
  number={10},
  pages={4780--4791},
  year={2023},
  publisher={IEEE}
}

@inproceedings{wang2023mcf,
  title={Mcf: Mutual correction framework for semi-supervised medical image segmentation},
  author={Wang, Yongchao and Xiao, Bin and Bi, Xiuli and Li, Weisheng and Gao, Xinbo},
  booktitle={Proceedings of the IEEE/CVF conference on computer vision and pattern recognition},
  pages={15651--15660},
  year={2023}
}

@article{xu2021asymmetric,
  title={Asymmetric multi-task attention network for prostate bed segmentation in computed tomography images},
  author={Xu, Xuanang and Lian, Chunfeng and Wang, Shuai and Zhu, Tong and Chen, Ronald C and Wang, Andrew Z and Royce, Trevor J and Yap, Pew-Thian and Shen, Dinggang and Lian, Jun},
  journal={Medical image analysis},
  volume={72},
  pages={102116},
  year={2021},
  publisher={Elsevier}
}

@article{malhotra2022multi,
  title={Multi-task driven explainable diagnosis of COVID-19 using chest X-ray images},
  author={Malhotra, Aakarsh and Mittal, Surbhi and Majumdar, Puspita and Chhabra, Saheb and Thakral, Kartik and Vatsa, Mayank and Singh, Richa and Chaudhury, Santanu and Pudrod, Ashwin and Agrawal, Anjali},
  journal={Pattern recognition},
  volume={122},
  pages={108243},
  year={2022},
  publisher={Elsevier}
}

@article{zhou2020benchmark,
  title={A benchmark for studying diabetic retinopathy: segmentation, grading, and transferability},
  author={Zhou, Yi and Wang, Boyang and Huang, Lei and Cui, Shanshan and Shao, Ling},
  journal={IEEE transactions on medical imaging},
  volume={40},
  number={3},
  pages={818--828},
  year={2020},
  publisher={IEEE}
}

@inproceedings{tang2019nodulenet,
  title={Nodulenet: Decoupled false positive reduction for pulmonary nodule detection and segmentation},
  author={Tang, Hao and Zhang, Chupeng and Xie, Xiaohui},
  booktitle={Medical Image Computing and Computer Assisted Intervention--MICCAI 2019: 22nd International Conference, Shenzhen, China, October 13--17, 2019, Proceedings, Part VI 22},
  pages={266--274},
  year={2019},
  organization={Springer}
}

@article{tomar2021self,
  title={Self-attentive spatial adaptive normalization for cross-modality domain adaptation},
  author={Tomar, Devavrat and Lortkipanidze, Manana and Vray, Guillaume and Bozorgtabar, Behzad and Thiran, Jean-Philippe},
  journal={IEEE transactions on medical imaging},
  volume={40},
  number={10},
  pages={2926--2938},
  year={2021},
  publisher={IEEE}
}

@article{zhou2020hi,
  title={Hi-net: hybrid-fusion network for multi-modal MR image synthesis},
  author={Zhou, Tao and Fu, Huazhu and Chen, Geng and Shen, Jianbing and Shao, Ling},
  journal={IEEE transactions on medical imaging},
  volume={39},
  number={9},
  pages={2772--2781},
  year={2020},
  publisher={IEEE}
}

@article{hamghalam2020high,
  title={High tissue contrast image synthesis via multistage attention-GAN: application to segmenting brain MR scans},
  author={Hamghalam, Mohammad and Wang, Tianfu and Lei, Baiying},
  journal={Neural Networks},
  volume={132},
  pages={43--52},
  year={2020},
  publisher={Elsevier}
}

@article{yang2021hybrid,
  title={A hybrid deep segmentation network for fundus vessels via deep-learning framework},
  author={Yang, Lei and Wang, Huaixin and Zeng, Qingshan and Liu, Yanhong and Bian, Guibin},
  journal={Neurocomputing},
  volume={448},
  pages={168--178},
  year={2021},
  publisher={Elsevier}
}

@article{wang2020automatic,
  title={Automatic ischemic stroke lesion segmentation from computed tomography perfusion images by image synthesis and attention-based deep neural networks},
  author={Wang, Guotai and Song, Tao and Dong, Qiang and Cui, Mei and Huang, Ning and Zhang, Shaoting},
  journal={Medical Image Analysis},
  volume={65},
  pages={101787},
  year={2020},
  publisher={Elsevier}
}

@inproceedings{zhao2022cross,
  title={Cross-level contrastive learning and consistency constraint for semi-supervised medical image segmentation},
  author={Zhao, Xinkai and Fang, Chaowei and Fan, De-Jun and Lin, Xutao and Gao, Feng and Li, Guanbin},
  booktitle={2022 IEEE 19th International Symposium on Biomedical Imaging (ISBI)},
  pages={1--5},
  year={2022},
  organization={IEEE}
}

@inproceedings{hou2022semi,
  title={Semi-supervised semantic segmentation of vessel images using leaking perturbations},
  author={Hou, Jinyong and Ding, Xuejie and Deng, Jeremiah D},
  booktitle={Proceedings of the IEEE/CVF winter conference on applications of computer vision},
  pages={2625--2634},
  year={2022}
}

@article{wang2024boundary,
  title={Boundary-Aware Prototype in Semi-Supervised Medical Image Segmentation},
  author={Wang, YongChao and Xiao, Bin and Bi, Xiuli and Li, Weisheng and Gao, Xinbo},
  journal={IEEE Transactions on Image Processing},
  year={2024},
  publisher={IEEE}
}

@article{jiao2024learning,
  title={Learning with limited annotations: a survey on deep semi-supervised learning for medical image segmentation},
  author={Jiao, Rushi and Zhang, Yichi and Ding, Le and Xue, Bingsen and Zhang, Jicong and Cai, Rong and Jin, Cheng},
  journal={Computers in Biology and Medicine},
  volume={169},
  pages={107840},
  year={2024},
  publisher={Elsevier}
}

@inproceedings{ma2020distance,
  title={How distance transform maps boost segmentation CNNs: an empirical study},
  author={Ma, Jun and Wei, Zhan and Zhang, Yiwen and Wang, Yixin and Lv, Rongfei and Zhu, Cheng and Gaoxiang, Chen and Liu, Jianan and Peng, Chao and Wang, Lei and others},
  booktitle={Medical Imaging with Deep Learning},
  pages={479--492},
  year={2020},
  organization={PMLR}
}

@inproceedings{li2020shape,
  title={Shape-aware semi-supervised 3D semantic segmentation for medical images},
  author={Li, Shuailin and Zhang, Chuyu and He, Xuming},
  booktitle={Medical Image Computing and Computer Assisted Intervention--MICCAI 2020: 23rd International Conference, Lima, Peru, October 4--8, 2020, Proceedings, Part I 23},
  pages={552--561},
  year={2020},
  organization={Springer}
}

@article{luo2022semi,
  title={Semi-supervised medical image segmentation via uncertainty rectified pyramid consistency},
  author={Luo, Xiangde and Wang, Guotai and Liao, Wenjun and Chen, Jieneng and Song, Tao and Chen, Yinan and Zhang, Shichuan and Metaxas, Dimitris N and Zhang, Shaoting},
  journal={Medical Image Analysis},
  volume={80},
  pages={102517},
  year={2022},
  publisher={Elsevier}
}

@article{wang2021self,
  title={Self-paced and self-consistent co-training for semi-supervised image segmentation},
  author={Wang, Ping and Peng, Jizong and Pedersoli, Marco and Zhou, Yuanfeng and Zhang, Caiming and Desrosiers, Christian},
  journal={Medical Image Analysis},
  volume={73},
  pages={102146},
  year={2021},
  publisher={Elsevier}
}

@article{li2023semi,
  title={Semi-supervised detection model based on adaptive ensemble learning for medical images},
  author={Li, Jingchen and Shi, Haobin and Chen, Wenbai and Liu, Naijun and Hwang, Kao-Shing},
  journal={IEEE Transactions on Neural Networks and Learning Systems},
  year={2023},
  publisher={IEEE}
}

@article{wang2022semi,
  title={Semi-supervised medical image segmentation via a tripled-uncertainty guided mean teacher model with contrastive learning},
  author={Wang, Kaiping and Zhan, Bo and Zu, Chen and Wu, Xi and Zhou, Jiliu and Zhou, Luping and Wang, Yan},
  journal={Medical Image Analysis},
  volume={79},
  pages={102447},
  year={2022},
  publisher={Elsevier}
}

@inproceedings{bai2023bidirectional,
  title={Bidirectional copy-paste for semi-supervised medical image segmentation},
  author={Bai, Yunhao and Chen, Duowen and Li, Qingli and Shen, Wei and Wang, Yan},
  booktitle={Proceedings of the IEEE/CVF conference on computer vision and pattern recognition},
  pages={11514--11524},
  year={2023}
}

@article{han2022effective,
  title={An effective semi-supervised approach for liver CT image segmentation},
  author={Han, Kai and Liu, Lu and Song, Yuqing and Liu, Yi and Qiu, Chengjian and Tang, Yangyang and Teng, Qiaoying and Liu, Zhe},
  journal={IEEE Journal of Biomedical and Health Informatics},
  volume={26},
  number={8},
  pages={3999--4007},
  year={2022},
  publisher={IEEE}
}

@article{wu2023exploring,
  title={Exploring feature representation learning for semi-supervised medical image segmentation},
  author={Wu, Huimin and Li, Xiaomeng and Cheng, Kwang-Ting},
  journal={IEEE Transactions on Neural Networks and Learning Systems},
  year={2023},
  publisher={IEEE}
}

@article{shi2021inconsistency,
  title={Inconsistency-aware uncertainty estimation for semi-supervised medical image segmentation},
  author={Shi, Yinghuan and Zhang, Jian and Ling, Tong and Lu, Jiwen and Zheng, Yefeng and Yu, Qian and Qi, Lei and Gao, Yang},
  journal={IEEE transactions on medical imaging},
  volume={41},
  number={3},
  pages={608--620},
  year={2021},
  publisher={IEEE}
}

@inproceedings{miao2023caussl,
  title={Caussl: Causality-inspired semi-supervised learning for medical image segmentation},
  author={Miao, Juzheng and Chen, Cheng and Liu, Furui and Wei, Hao and Heng, Pheng-Ann},
  booktitle={Proceedings of the IEEE/CVF International Conference on Computer Vision},
  pages={21426--21437},
  year={2023}
}
\end{document}